\documentclass[lettersize,journal]{IEEEtran}
\usepackage{amsmath,amsfonts}
\usepackage{algorithmic}
\usepackage{algorithm}
\usepackage{array}
\usepackage[caption=false,font=normalsize,labelfont=sf,textfont=sf]{subfig}
\usepackage{textcomp}
\usepackage{stfloats}
\usepackage{xurl}
\usepackage{verbatim}
\usepackage{graphicx}
\usepackage{cite}
\usepackage{booktabs}
\usepackage{xcolor}
\usepackage{orcidlink}

\begin{document}

\title{Efficient Spatio-Temporal Vegetation Pixel Classification with\\ Vision Transformers}

\author{
\IEEEauthorblockN{Alan Gomes\orcidlink{0009-0005-6885-113X},
Anderson Gonçalves\orcidlink{0009-0003-7551-388X}, 
Samuel Felipe dos Santos\orcidlink{0000-0001-6061-5582}, 
Nathan Felipe Alves\orcidlink{0000-0002-9710-6332}, 
Magna Soelma Beserra de Moura\orcidlink{0000-0002-2844-1399}, 
Bruna de Costa Alberton\orcidlink{0000-0003-4835-8389}, 
Leonor Patricia C. Morellato\orcidlink{0000-0001-5265-8988}, 
Ricardo da Silva Torres\orcidlink{0000-0001-9772-263X}, and 
Jurandy Almeida\orcidlink{0000-0002-4998-6996}} 
\thanks{Alan Gomes, Anderson Gonçalves, Samuel Felipe dos Santos, and Jurandy Almeida are with the Department of Computing, Federal University of São Carlos (UFSCar), Sorocaba, 18052-780, Brazil (e-mail: \{alan.gomes, andersoncsg\}@estudante.ufscar.br and \{samuel.felipe, jurandy.almeida\}@ufscar.br).}
\thanks{Nathan Felipe Alves, Bruna de Costa Alberton, and Leonor Patricia C. Morellato are with the Center for Research on Biodiversity Dynamics and Climate Change and Department of Biodiversity, Bioscience Institute, São Paulo State University (UNESP), Rio Claro, 13506-900, Brazil (e-mail: \{nathan.felipe, bruna.alberton, patricia.morellato\}@unesp.br).}
\thanks{Magna Soelma Beserra de Moura is with the Brazilian Agricultural Research Corporation (EMBRAPA), Fortaleza, 60511-110, Brazil (e-mail: magna.moura@embrapa.br).}
\thanks{Ricardo da Silva Torres is with the Wageningen University \& Research (WUR), Wageningen, 6708PB, Netherlands (e-mail: ricardo.dasilvatorres@wur.nl).}
}



\maketitle

\begin{abstract}
Plant phenology—the study of recurrent life cycle events—is essential for understanding ecosystem dynamics and their responses to climate change impacts.
While Unmanned Aerial Vehicles (UAVs) and near-surface cameras enable high-resolution monitoring, identifying plant species across time remains computationally challenging.
State-of-the-art approaches, specifically Multi-Temporal Convolutional Networks~(CNNs), rely on rigid multi-branch architectures that scale poorly with longer time series and require large spatial context windows.
In this paper, we present an extensive study on optimizing Vision Transformers (ViTs) for efficient spatio-temporal vegetation pixel classification.
We conducted a comprehensive ablation study analyzing seven key design dimensions, including: (\textit{i}) data normalization; (\textit{ii}) spectral arrangement; (\textit{iii}) boundary handling; (\textit{iv}) {spatial} context window {shape and} size; (\textit{v}) tokenization strategies; (\textit{vi}) positional encoding; and (\textit{vii}) feature aggregation strategies.
Our method was evaluated on two datasets from the Brazilian Cerrado biome, Serra do Cipó (aerial imagery) and Itirapina (near-surface imagery). 
Experimental results demonstrate that our ViT approach offers a substantial improvement in computational efficiency while maintaining competitive classification performance.
Notably, our ViT reduces Floating Point Operations (FLOPs) by an order of magnitude and maintains constant parameter complexity regardless of the time series length, whereas the CNN baseline scales linearly. 
Our findings confirm that ViTs are a robust, scalable solution for resource-constrained phenological monitoring systems.
\end{abstract}


\section{Introduction}
\IEEEPARstart{W}{ith} the acceleration of global climate warming, there is a growing urgency for technologies capable of monitoring environmental changes and their impacts on biotic components. Developing systems capable of extracting actionable knowledge from environmental data is crucial for supporting environmental management policies and maintaining ecosystem balance~\cite{Clark2007,Torres2006,TPAMI_2016_Dosovitskiy}.

Plant phenology is the study of recurrent life cycle events such as leaf flushing, flowering, and fruiting. These transitions serve as a sensitive and reliable indicator of underlying ecosystem dynamics~\cite{Parmesan2003,Rosenzweig2008,Walther2004,Walther2002}. 
Phenological patterns are directly influenced by climate, hydrology, and soil conditions, making them vital metrics for tracking climate change~\cite{gong2024satellite}.
Specifically, leaf phenology (e.g., budburst and senescence) defines the growing season and controls essential ecosystem processes such as carbon cycling and water regulation~\cite{Aide1988,Morisette2009,Alberton2017,alberton2023relationship}.

\IEEEpubidadjcol

Traditionally, phenological data was gathered through direct, labor-intensive field observation. 
However, remote sensing strategies have revolutionized this field by enabling continuous, large-scale monitoring without direct contact~\cite{gong2024satellite}. 
While satellites provide global coverage, they often lack the spatial resolution required to identify individual species in diverse tropical ecosystems. 
To bridge this gap, aerial and near-surface technologies, such as Unmanned Aerial Vehicles (UAVs) and phenological towers, have emerged. They are capable of capturing high-resolution image time series with detailed spectral information~\cite{li2024review}.

A fundamental challenge in remote phenology lies in the accurate localization and identification of plant individuals within these dense image series~\cite{richardson2009near}.
This is particularly difficult in tropical vegetations like the Brazilian Cerrado, where numerous species co-exist in tightly packed communities with overlapping canopies~\cite{ECOI_2014_Alberton}.
Fortunately, while these species may look spatially similar, they often possess distinct phenological traits, flowering or losing leaves at different times of the year. Consequently, their temporal profile act as a unique fingerprint~\cite{ECOI_2014_Alberton,Morellato2013}. 
To accurately identify vegetation in these environments, a method must therefore effectively combine spatial cues with these discriminative temporal patterns~\cite{nogueira2019spatio}.

This requirement defines the task of spatio-temporal vegetation pixel classification, which involves analyzing a time series of spatially aligned images in order to identify plant species based on the evolution of their visual patterns over time.
To do this, machine learning models are trained to assign a species label to every pixel based on its phenological evolution, taking into account the pixel itself and its neighboring pixels~\cite{nogueira2019spatio}.

Early approaches to this problem relied on hand-crafted features and standard classifiers like SVMs~\cite{almeida2014applying,almeida2016unsupervised,almeida2016phenological,almeida2015deriving,faria2016time,faria2016fusion}.
More recently, deep learning has enabled the ability to learn intricate features and complex classifiers directly from raw data in an end-to-end manner.
State-of-the-art approaches, such as the Multi-Temporal Convolutional Neural Network~(CNN) proposed by Nogueira~et~al.~\cite{nogueira2019spatio}, have achieved high accuracy by employing multi-branch architectures where each timestamp is processed by a dedicated convolutional stream. 
While effective, this architecture suffers from structural rigidity and computational inefficiency, since it relies on computationally expensive spatial convolutions over large context windows (25x25 patches) and processes time via channel concatenation or independent branches, causing the model’s parameter count to grow linearly with the time series length.

In this context, Vision Transformers (ViTs) offer a more efficient and scalable alternative. 
Unlike CNNs, which process fixed grids, Transformers treat time series as sequences of tokens. 
This allows for flexible modeling of temporal dependencies via self-attention mechanisms~\cite{li2024review}. 
Although ViTs have achieved state-of-the-art results in satellite remote sensing~\cite{li2025agrifm}, their application to high-resolution phenological time series remains largely unexplored.

In this paper, we bridge this gap by presenting an extensive study on optimizing ViTs for efficient spatio-temporal vegetation pixel classification {in aerial and near-surface tower imagery.} 
We hypothesize that by prioritizing the temporal sequence via self-attention, we can drastically reduce the reliance on spatial context, leading to a more efficient and scalable architecture.
{We focus our evaluation on the unique challenges presented by the tropical environments of the Brazilian Cerrado.}
%
Our main contributions are:

\begin{itemize}
    \item We propose and evaluate a ViT framework specifically adapted for the spatio-temporal vegetation pixel classification in high-resolution image time series from phenological monitoring.

    \item We conduct a comprehensive ablation study analyzing seven key design choices—including 
(1) the use of data normalization;
(2) the arrangement of color channels within pixel groups;
(3) the policy for handling pixels outside the mask boundaries;
(4) the influence of the spatial context window {shape and} size;
(5) the comparison between spatial and temporal tokens;
(6) the use of positional encoding; and
(7) the feature aggregation strategy—to determine the optimal architecture for phenological data.

    \item We demonstrate that our ViT architecture yields a classification performance comparable or better than the state-of-the-art Multi-Temporal CNN~\cite{nogueira2019spatio} while offering an order-of-magnitude reduction in Floating Point Operations (FLOPs). This is achieved by maintaining constant parameter complexity regardless of time series length, making it ideal for resource-constrained environments.
\end{itemize}

The remainder of this paper is organized as follows.
Section~\ref{sec:related_works} presents related work.
Section~\ref{sec:problem_definition} defines the spatio-temporal vegetation pixel classification problem.
Section~\ref{sec:proposed_methodologies} introduces our proposed framework.
Section~\ref{sec:experimental_protocol} describes the experimental protocol.
Section~\ref{sec:results} reports and analyzes our experimental results.
Finally, Section~\ref{sec:conclusion} offers conclusions and directions for future research.

\section{Related Work}
\label{sec:related_works}
Early approaches for plant species identification and phenological monitoring relied on hand-crafted features (e.g., color histograms, texture analysis) and standard classifiers like SVMs~\cite{almeida2014applying,almeida2016unsupervised,almeida2016phenological,almeida2015deriving,faria2016time,faria2016fusion}. 
While effective for small datasets, these methods struggle to generalize across the high intraclass variance typical of tropical vegetation.

The introduction of CNNs marked a paradigm shift, allowing for the learning of data-driven features directly from raw images. 
CNNs have since become the standard for various agricultural and environmental tasks, including crop disease detection and land cover mapping~\cite{barbedo2025review}. 
However, standard CNNs are inherently designed for spatial feature extraction in static images. 
To capture phenology—which is defined by change over time—architectures must be adapted to process the temporal dimension effectively.

To address the temporal dynamics of vegetation, Nogueira~et~al.~\cite{nogueira2019spatio} introduced the Multi-Temporal CNN, a pixel-wise classification framework that employs separate convolutional branches for each timestamp in the series, fusing them at later stages to capture phenological profiles. 
While effective at capturing local texture and achieving high accuracy, it is structurally rigid and computationally expensive, with parameter counts that grow linearly with the duration of the monitoring campaign.

The emergence of ViTs~\cite{dosovitskiy2020image} has challenged the dominance of CNNs by introducing self-attention mechanisms that capture long-range dependencies and global context. 
However, most Transformer-based remote sensing research focuses on satellite imagery, where pixel resolution is low (10-30m)~\cite{victor2025satelliteimagery}.
For instance,  Li~et~al.~\cite{li2025agrifm} proposed AgriFM, which employs a Video Swin Transformer to extract hierarchical features from multi-source temporal satellite data (Sentinel, Landsat, MODIS).
While Transformers excel at satellite scales, their application to high-resolution phenology remains largely unexplored. 
Unlike satellite data, aerial and near-surface imagery offers fine-grained detail required for identifying individual plants in diverse tropical communities~\cite{ECOI_2014_Alberton}.

To date, there is limited work systematically optimizing ViTs for the specific constraints of high-resolution phenological time series. 
This work addresses this gap, investigating how ViTs can be streamlined—specifically regarding tokenization and input sparsity—to offer a scalable alternative to multi-branch CNNs.
We overcome the limitations of CNN-based methods by proposing a ViT model that captures temporal evolution with minimal spatial context, offering an accurate and computationally scalable solution for resource-constrained environments.

\section{Spatio-temporal Vegetation Pixel Classification}
\label{sec:problem_definition}
In this work, we address the problem of spatio-temporal vegetation pixel classification as defined by Nogueira~et~al.~\cite{nogueira2019spatio}.
Formally, let $\mathcal{S} = \{ I_{t_1}, I_{t_2}, \dots, I_{t_M} \}$ be a sequence of $M$ high-resolution images captured over a specific geographic region at timestamps $T = \left\{ t_1, t_2, \dots, t_M \right\}$.
We assume that all images in $\mathcal{S}$ are spatially co-registered, such that a pixel coordinate $p=(x, y)$ corresponds to the same physical location across all timestamps $t_i \in T$.

For any given pixel location $p=(x, y)$, we can extract a feature vector $\mathbf{x}_{p} = \left[ I_{t_1}(p), I_{t_2}(p), \dots, I_{t_M}(p) \right]$ that encapsulates its spectral evolution, where $I_{t_i}(p) = \left\{ R_{t_i}^p, G_{t_i}^p, R_{t_i}^p \right\}$ represents the RGB values of the pixel $p$ at time $t_i$.

However, in high-resolution imagery, a single pixel often contains noise or lacks sufficient texture. 
To mitigate this, the input is expanded to include a spatial context window—a patch of size $k \times k$ centered at the pixel $p=(x, y)$.
Let $\mathcal{N}_r(p) = \{ q \in \mathbb{Z}^2 \vert d(p, q) \leq r \}$ be the set of spatial coordinates in the neighborhood of a given pixel $p=(x, y)$ within an arbitrary radius $r = \left\lfloor \frac{k}{2} \right\rfloor$, where $q=(i, j)$ is a pixel coordinate and $d(p, q)$ is the distance between pixels $p$ and $q$ using a given metric (e.g., Manhattan, Euclidean, Chebyshev).
We define the spatio-temporal volume $\mathbf{X}_{p} = \left\{ \mathbf{x}_{q} \left| q \in \mathcal{N}_r(p) \right. \right\}$ as the aggregation of the temporal vectors of all pixels within the spatial neighborhood.

The objective is to learn a mapping function $f : \mathcal{X} \rightarrow \mathcal{Y}$ that predicts the species class $y_p \in \mathcal{Y}$ for the target pixel $p=(x, y)$ based on its spatio-temporal volume $\mathbf{X}_{p}$.
The core challenge of this task lies in the heterogeneity of these volumes. 
The model must be invariant to illumination changes across months and robust to missing data (e.g., cloud cover), while simultaneously learning to distinguish subtle phenological traits—such as the specific timing of leaf loss or flowering—that differentiate typically green species in complex tropical communities.

\section{Vision Transformer for Phenological Time Series}
\label{sec:proposed_methodologies}
To address the limitations of prior CNN architectures, we adopt a Vision Transformer (ViT) as the core modeling engine for spatio-temporal vegetation pixel classification.
Unlike CNNs that process data via fixed local neighborhoods, ViTs process data as sequences of tokens, employing self-attention mechanisms to capture global dependencies.
This allows the model to flexibly weigh the importance of different timestamps or spatial neighbors without being constrained by a fixed kernel size.

In this section, we first detail the generic ViT architecture adapted for this task, and then define the seven-dimensional design space we explored to optimize this architecture for high-resolution phenological data.

\begin{figure*}[!htb]
    \centering
    \includegraphics[width=1\linewidth]{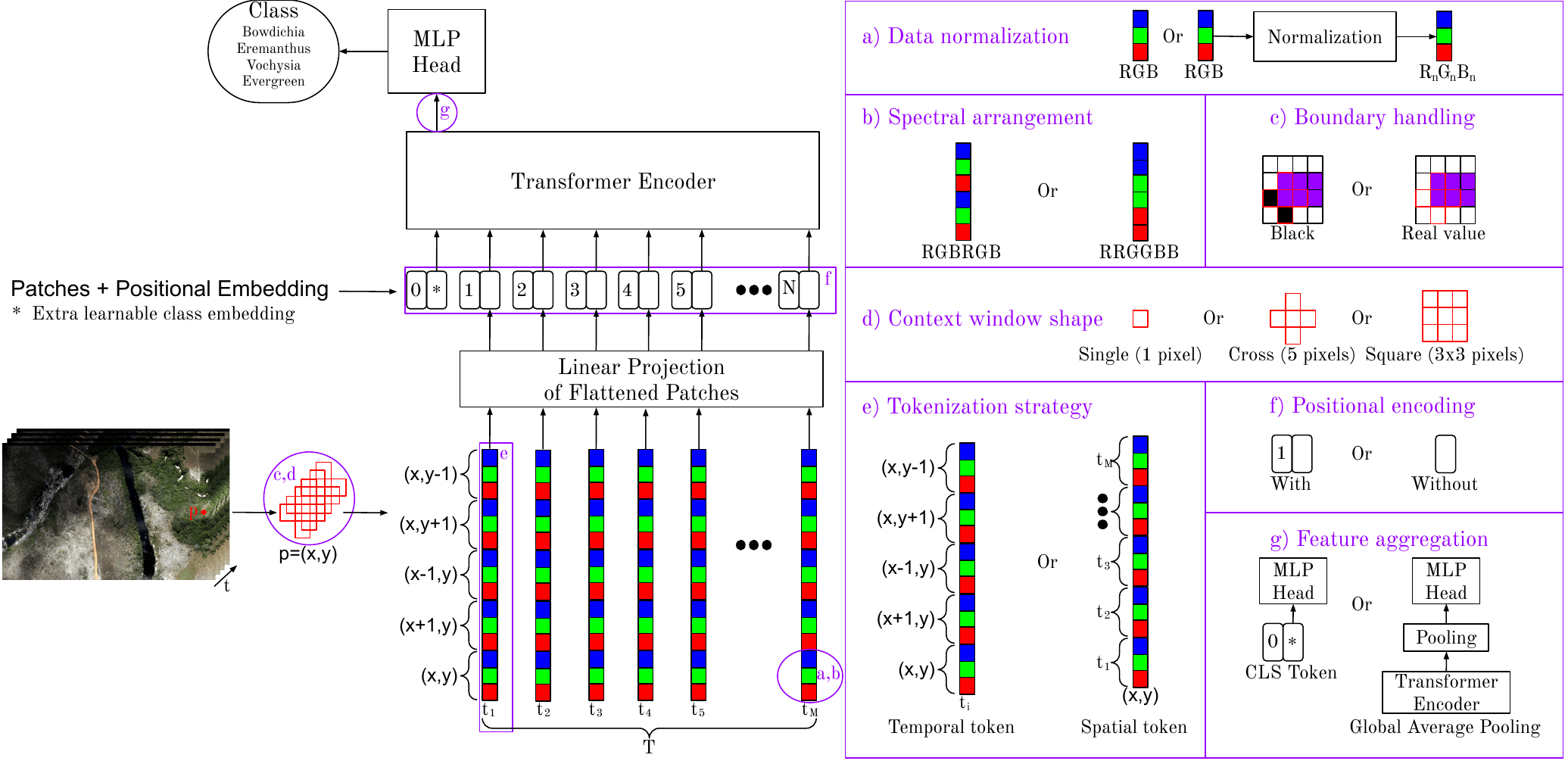}
    \caption{Overview of our method for searching for an optimal setting for applying a ViT architecture to the spatio-temporal vegetation pixel classification task. We investigated the following settings: (a) data normalization; (b) color channel arrangement; (c) policies for pixels outside boundaries; (d) {spatial} context window shape; (e) tokenization strategies; (f) positional encoding; and (g) feature aggregation strategies.}
    \label{fig:overview}
\end{figure*}

\subsection{Network Architecture}
Our model follows the standard ViT encoder architecture~\cite{dosovitskiy2020image}, adapted to process 1D sequences of spatio-temporal data rather than 2D image patches, as illustrated in Figure~\ref{fig:overview}. 
First, the spatio-temporal volume $\mathbf{X}_{p} \in \mathbb{R}^{M \times k \times k \times 3}$ for each pixel $p=(x, y)$ is flattened into a sequence of $N$ vectors (tokens) $\mathbf{X}_{p}^{flatten} \in \mathbb{R}^{N \times D_{in}}$.
Then, this sequence is linearly projected into a latent embedding space of dimension $D$ using a learnable projection matrix $\mathbf{E} \in \mathbb{R}^{D_{in} \times D}$. 
To retain sequence order information, a learnable positional embedding $\mathbf{E}_{pos} \in \mathbb{R}^{N \times D}$ is added. 
If a class token is used, a learnable vector $\mathbf{z}_{cls}$ is prepended to the sequence:
\begin{displaymath}
    \mathbf{Z}_0 = \left[ \mathbf{z}_0^{cls} ; \mathbf{X}_{p}^{flatten} \mathbf{E} \right] + \mathbf{E}_{pos}
\end{displaymath}

Next, the sequence $\mathbf{Z}_0$ is processed by $L$ layers of Transformer blocks. Each block consists of Multi-Head Self-Attention~(MSA) and a Multi-Layer Perceptron~(MLP), with Layer Normalization~(LN) applied before each block and residual connections added after:
\begin{align}
    \mathbf{Z}'_l &= \text{MSA}(\text{LN}(\mathbf{Z}_{l-1})) + \mathbf{Z}_{l-1} \\
    \mathbf{Z}_l  &= \text{MLP}(\text{LN}(\mathbf{Z}'_l)) + \mathbf{Z}'_l
\end{align}

Thereafter, the output of the final layer $\mathbf{Z}_L$ is aggregated into a single feature vector $\mathbf{y}_p$.
Depending on the aggregation strategy, $\mathbf{y}_p$ is derived either from the class token state, $\mathbf{y}_p = \mathbf{Z}_L^0$; or the global average of the sequence, $\mathbf{y}_p = \frac{1}{N} \sum_{i=1}^N \mathbf{Z}_L^i$.
Finally, the final representation $\mathbf{y}_p$ is processed to produce the class probability distribution $y_p$: 
\begin{displaymath}
    y_p = \text{Softmax}(\text{MLP}_{head}(\mathbf{y}_p)).
\end{displaymath}

\subsection{Design Space Exploration}
To determine the optimal strategy for adapting ViTs to phenological time series, we investigate a design space covering seven key dimensions, as illustrated in Figure~\ref{fig:overview}. 
These configurations, labeled (a) through (g) in the diagram, govern how phenological data is handled from raw input to final classification.

\subsubsection{Data Normalization}
Phenological traits are often defined by subtle changes in absolute reflectance (e.g., the brightening of a canopy during flowering). 
We investigate whether standard normalization aids or hinders this signal (see Figure~\ref{fig:overview}(a)):
\begin{itemize}
    \item \textbf{Raw Data:} The pixel intensity $I_{t_i}(p) = \left\{ R_{t_i}^p, G_{t_i}^p, B_{t_i}^p \right\}$ is used directly, preserving the absolute radiometric information and seasonal lighting variations.

    \item \textbf{Normalized Data:} Each channel is normalized by the total intensity of the pixel, such that $I_{t_i}(p) = \left\{ \frac{R_{t_i}^p}{R_{t_i}^p + G_{t_i}^p + B_{t_i}^p}, \frac{G_{t_i}^p}{R_{t_i}^p + G_{t_i}^p + B_{t_i}^p}, \frac{B_{t_i}^p}{R_{t_i}^p + G_{t_i}^p + B_{t_i}^p} \right\}$, focusing the model on chromaticity (pure color) rather than intensity.
\end{itemize}

\subsubsection{Spectral Arrangement}
We analyze whether the internal ordering of spectral bands within a token influences the initial feature extraction (see Figure~\ref{fig:overview}(b)):
\begin{itemize}
    \item \textbf{Natural Order (rgbrgb):} Preserves the temporal sequence of channels, i.e.,
    \begin{displaymath}
        \mathbf{x}_p = \left[ R_{t_1}^p, G_{t_1}^p, B_{t_1}^p, R_{t_2}^p, G_{t_2}^p, B_{t_2}^p, \dots, R_{t_M}^p, G_{t_M}^p, B_{t_M}^p \right].    
    \end{displaymath}
    
    \item \textbf{Channel Grouping (rrggbb):} Groups identical channels across time, i.e.,
    \begin{displaymath}
        \mathbf{x}_p = \left[ R_{t_1}^p, \dots, R_{t_M}^p, G_{t_1}^p, \dots, G_{t_2}^p, B_{t_M}^p, \dots, B_{t_M}^p \right].
    \end{displaymath}

    This potentially helps the embedding layer correlate single-channel evolution before mixing spectral data.
    However, the effect of the input order is closely related to how the projection layer is implemented. For instance, we expect the impact of this variation to be minimal when a linear layer is used for the embedding layer.

\end{itemize}

\subsubsection{Boundary Handling}
During training, when extracting a context window for a pixel located on the edge of the labeled mask indicating all pixels belonging to an individual of a given plant species, neighboring pixels may fall outside the annotated region. 
We evaluate two approaches (see Figure~\ref{fig:overview}(c)):
\begin{itemize}
    \item \textbf{Black Padding:} Neighbor pixels outside the labeled mask are set to zero (black). This forces the model to rely solely on the valid portion of the neighborhood.
    \item \textbf{Real Value:} The actual pixel values of the external neighbors are included, potentially providing useful context despite the lack of a specific label.
\end{itemize}

\subsubsection{Context Window Shape}
To classify a pixel, it is usually necessary a minimal amount of context information about the other surrounding pixels, but there are different manners to define the shape of this context window (see Figure~\ref{fig:overview}(d)). 
We evaluate three configurations by varying the distance metric $d$ and radius $r$:
\begin{itemize}
    \item \textbf{Single (1 pixel):} Defined by setting $r=0$. The model processes only the target pixel $p=(x, y)$—i.e., $\mathcal{N}_0(p) = \{ (x, y) \}$. This represents the maximum possible efficiency {while having no contextual information}, reducing the input feature vector to a purely temporal sequence. 
    \item \textbf{Cross-context window (5 pixels):} Defined by using the Manhattan distance ($L_1$ norm) with $r=1$. This selects the target pixel $p=(x, y)$ and its four orthogonal neighbors (von Neumann neighborhood), capturing immediate adjacency without the diagonal elements—i.e., $\mathcal{N}_1(p) = \{ (x, y); (x+1, y); (x-1, y); (x, y+1); (x, y-1) \}$.
    \item \textbf{Square context window (3$\times$3 pixels):} Defined by using the Chebyshev distance ($L_\infty$ norm) with $r=1$. This selects the target pixel $p=(x, y)$ and all eight surrounding neighbors (Moore neighborhood), effectively forming a dense $3 \times 3$ grid ($k=3)$—i.e., $\mathcal{N}_1(p) = \{ (x, y); (x+1, y); (x-1, y); (x, y+1); (x, y-1); (x+1, y+1); (x+1, y-1); (x-1, y+1); (x-1, y-1) \}$.
\end{itemize}

\subsubsection{Tokenization Strategy}
This dimension determines the semantics of the sequence length $N$ and the input dimension $D_{in}$.
We assess two orthogonal methods for flattening the spatio-temporal volume into a sequence of tokens (see Figure~\ref{fig:overview}(e)):
\begin{itemize}
    \item \textbf{Temporal Token (T, S):} The sequence is ordered by timestamp—i.e., $N=M, D_{in}=k \cdot k \cdot 3$. Each token contains all spatial data for a single moment $t_i$. This allows the self-attention mechanism to model dependencies between different dates.
    \item \textbf{Spatial Token (S, T):} The sequence is ordered by spatial position—i.e., $N=k \cdot k, D_{in}=M \cdot 3$. Each token represents a single pixel location and contains its entire temporal history. This forces self-attention to model dependencies between spatial neighbors.
\end{itemize}

\subsubsection{Positional Encoding}
Since Transformers are permutation-invariant, we assess the effect of adding positional encoding $\mathbf{E}_{pos}$ to the input tokens to incorporate information regarding temporal order or spatial layout (see Figure~\ref{fig:overview}(f)):
\begin{itemize}
    \item \textbf{With Positional Encoding:} $\mathbf{E}_{pos}$ consists of learnable parameters, initialized randomly and optimized during training, to explicitly encode the sequence order (temporal or spatial).
    \item \textbf{Without Positional Encoding:} $\mathbf{E}_{pos}$ is omitted, forcing the model to rely solely on token content, assuming that the intrinsic phenological state is sufficient to characterize the sequence.
\end{itemize}

\subsubsection{Feature Aggregation}
Finally, we analyze strategies to aggregate the output sequence $\mathbf{Z}_L$ into the feature vector $\mathbf{y}_p$ passed to the classifier $\text{MLP}_{head}$ (see Figure~\ref{fig:overview}(g)):
\begin{itemize}
    \item \textbf{Class Token ([CLS]):} A learnable token $\mathbf{z}_0^{cls}$ is prepended to the sequence. Its corresponding output embedding serves as the aggregate representation for classification, such that $\mathbf{y}_p = \mathbf{Z}_L^0$.
    \item \textbf{Global Average Pooling (GAP):} The output embeddings of all tokens in the sequence are averaged to form the final representation—i.e., $\mathbf{y}_p = \frac{1}{N} \sum_{i=1}^N \mathbf{Z}_L^i$.
\end{itemize}

\section{Experimental Protocol}
\label{sec:experimental_protocol}
To evaluate our ViT framework and validate its efficiency against the state-of-the-art, we established a rigorous experimental protocol. 
This section describes the datasets, implementation details, and evaluation metrics.

\subsection{Datasets}
We utilized two high-resolution spatio-temporal datasets from the Brazilian Cerrado biome, representing distinct acquisition modalities (UAV-based vs. tower-based imagery) and vegetation profiles.

\subsubsection{Serra do Cipó (UAV-based Imagery)} 
\begin{figure}[!htb]
    \centering
    \includegraphics[width=0.45\linewidth]{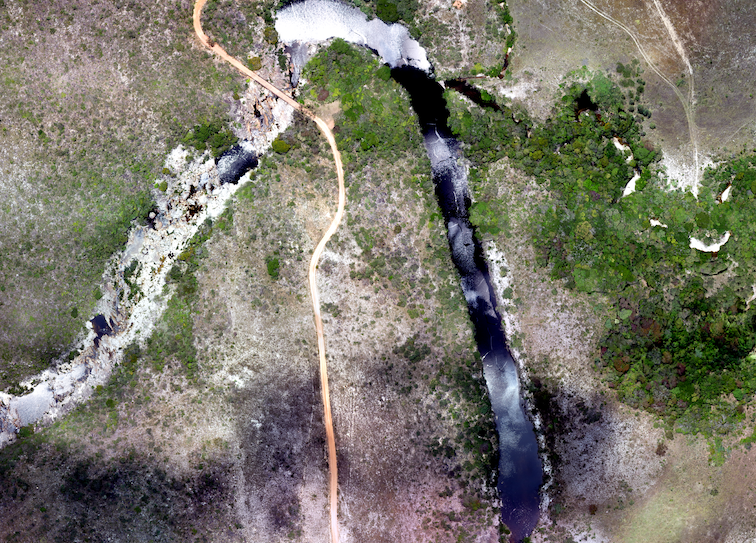}
    \includegraphics[width=0.45\linewidth]{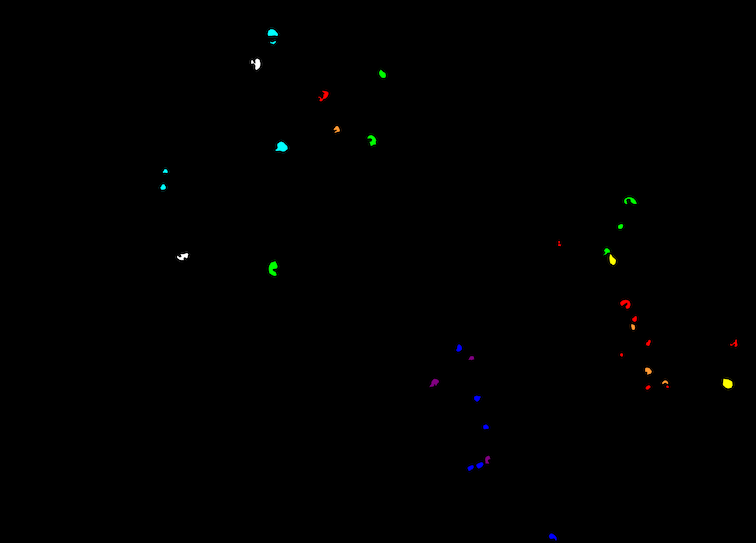}
    \caption{Sample RGB image from the Serra do Cipó dataset (left) and its ground-truth map (right). Following the protocol of Nogueira~et~al.~\cite{nogueira2019spatio}, the classes and their respective training and test splits are color-coded as: \textit{Bowdichia virgilioides} (red: train, orange: test); \textit{Eremanthus erythropappus} (blue: train, purple: test); \textit{Vochysia cinnamomea} (cyan: train, white: test); and a set of Evergreen species (green: train, yellow: test). Unclassified and background pixels are shown in black.}
    \label{fig:serra_do_cipo}
\end{figure}

This dataset consists of a time series of high-resolution aerial RGB images acquired over a \textit{Campo rupestre} (rocky grassland) vegetation in Santana do Riacho, Minas Gerais, Brazil~\cite{nogueira2019spatio}.
The images were captured using a Canon SX260 camera mounted on a fixed-wing UAV. 
The time series comprises 13 co-registered orthomosaics (6786$\times$9069 pixels) acquired monthly between February 2016 and February 2017.
Figure~\ref{fig:serra_do_cipo} presents a visual example of a specific timestamp alongside its corresponding ground-truth annotation.
The challenge in this dataset lies in the high biodiversity and spectral similarity of the vegetation. Annotations provided by expert botanists identify four distinct classes:(\textit{i}) \textit{Bowdichia virgilioides}, (\textit{ii}) \textit{Eremanthus erythropappus}, (\textit{iii}) \textit{Vochysia cinnamomea}, and (\textit{iv}) a functional group of Evergreen species.
As shown in Table~\ref{tab:serra_do_cipo}, the dataset exhibits significant class imbalance, with the Evergreen group accounting for the majority of annotated pixels.

\begin{table}[!htb]
    \centering
    \begin{tabular}{cc}
        \toprule
        \textbf{Classes} & \textbf{\#Pixels}  \\
        \hline
        \textit{Bowdichia virgilioides}   &  33,137 \\        
        \textit{Eremanthus erythropappus} &  31,250 \\
        \textit{Vochysia cinnamomea}      &  34,754 \\
        Evergreen species                 &  48,041 \\
        \hline
        Total                             & 147,182 \\
        \bottomrule
    \end{tabular}
    \caption{Distribution of annotated pixels per class in the Serra do Cipó dataset.}
    \label{tab:serra_do_cipo}
\end{table}

\subsubsection{Itirapina (Tower-based Imagery)}
\begin{figure}[!htb]
    \centering
    \includegraphics[width=0.45\linewidth]{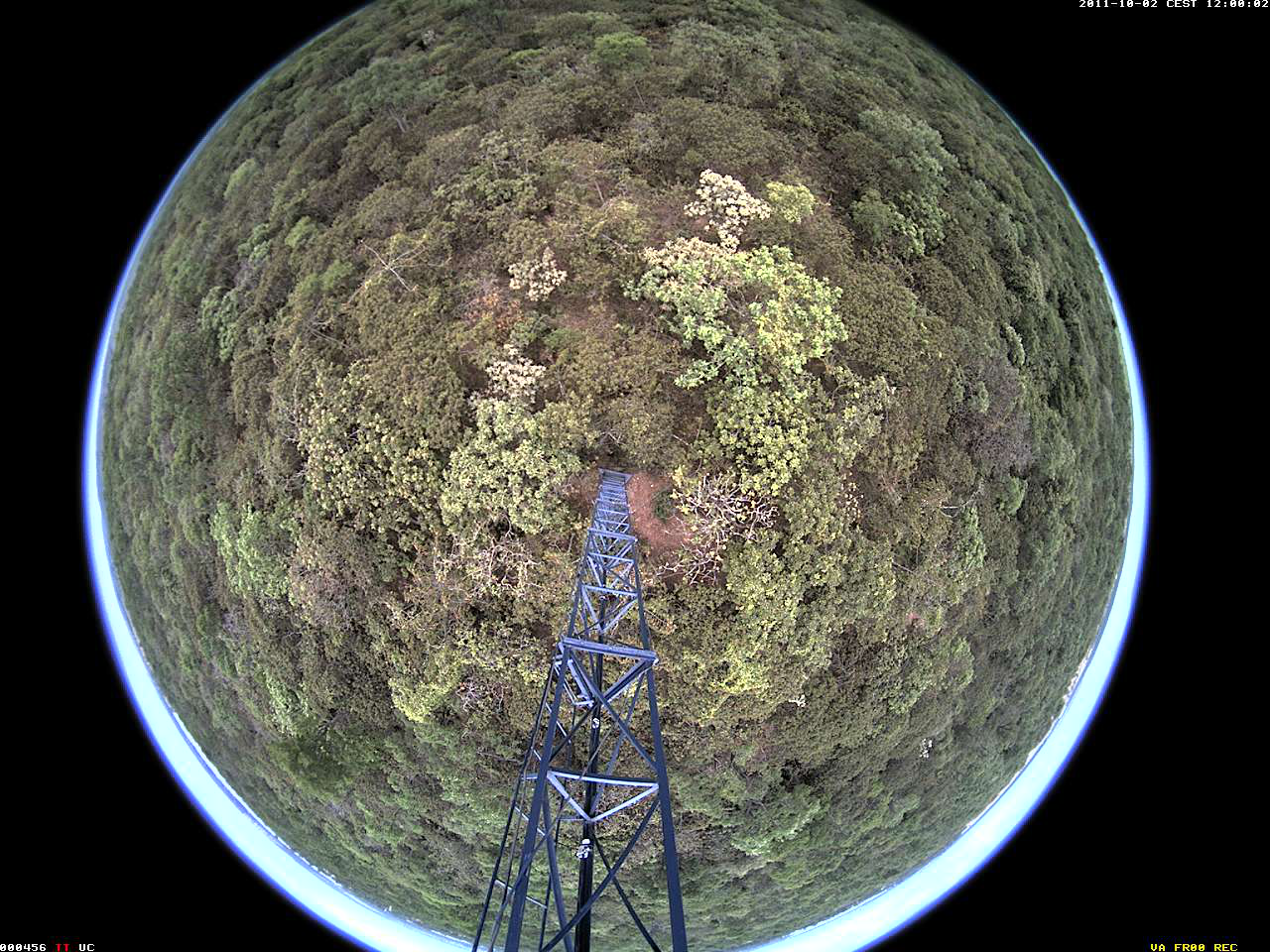}
    \includegraphics[width=0.45\linewidth]{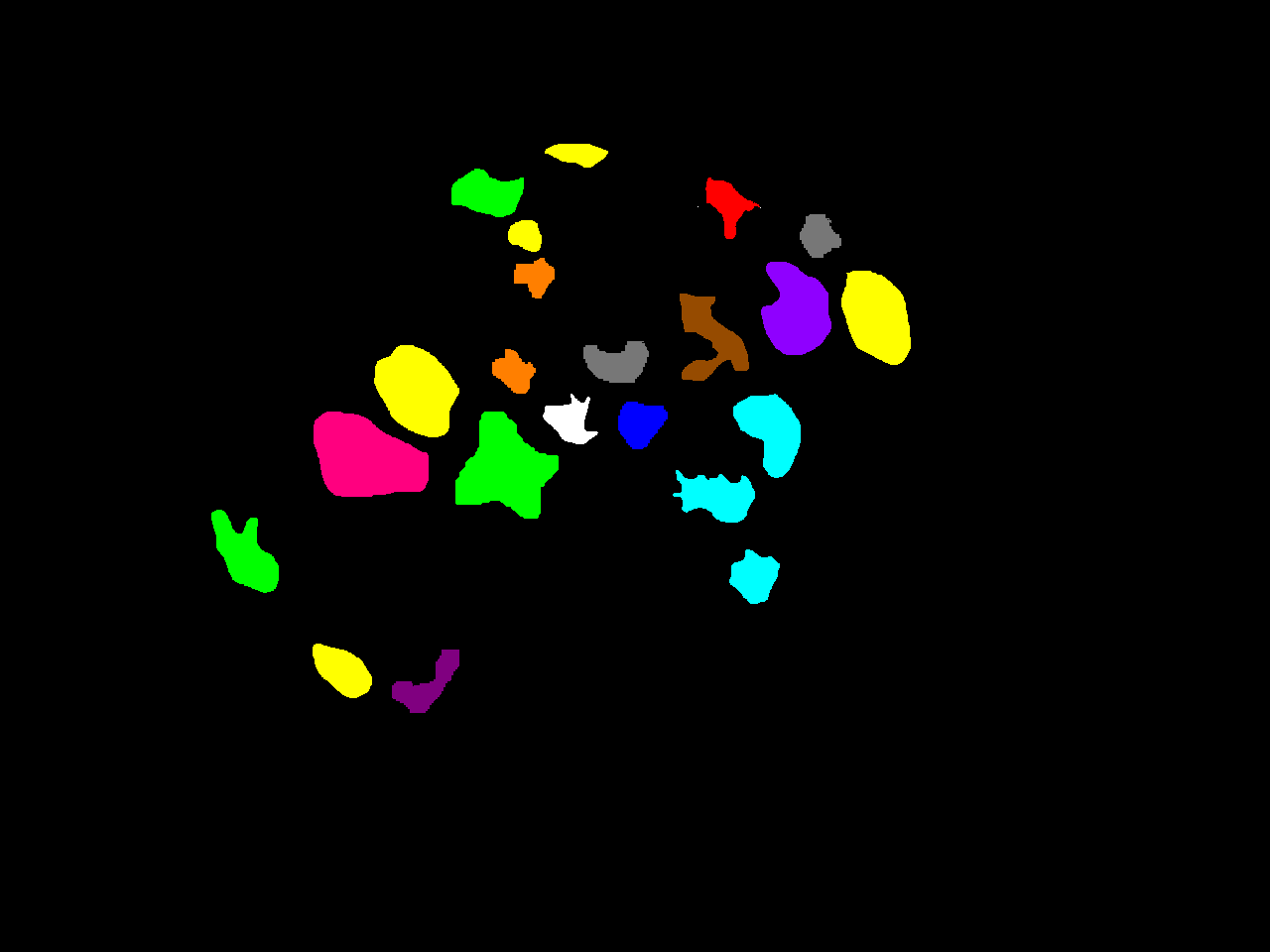}
    \caption{Sample RGB image from the Itirapina dataset (left) and its ground-truth map (right). The classes and their respective training and test splits are color-coded as: \textit{Aspidosperma tomentosum} (red: train, orange: test); \textit{Caryocar brasiliensis} (green: train, pink: test); \textit{Myrcia guianensis} (blue: train, white: test); \textit{Miconia rubiginosa} (yellow: train, gray: test); \textit{Pouteria ramiflora} (purple: train, brown: test); and \textit{Pouteria torta} (cyan: train, violet: test). Unclassified and background pixels are shown in black.}
    \label{fig:itirapina}
\end{figure}

The Itirapina dataset comprises high-resolution RGB images collected from a \textit{Cerrado sensu stricto} (savanna) reserve in Itirapina, São Paulo state, Brazil.
Data acquisition was performed by a near-surface phenological system featuring a digital camera mounted on an 18m tower.
While the system was configured to capture images hourly between 06:00 and 18:00 (UTC-3) with a resolution of 1280$\times$960 pixels, this study utilizes a specific time series of 36 timestamps recorded daily at 12:00 (noon) between August 29 and October 3, 2011.
Figure~\ref{fig:itirapina} displays a representative image from the dataset along with its ground-truth annotation. 
Crucially, we employ an updated version of this dataset containing significantly denser pixel annotations than the version used by Nogueira~et~al.~\cite{nogueira2019spatio}.
The distribution of annotated pixels per class is detailed in Table~\ref{tab:itirapina}.

\begin{table}[!htb]
    \centering
    \begin{tabular}{cc}
        \toprule
        \textbf{Classes} & \textbf{\#Pixels}  \\
        \hline
        \textit{Aspidosperma tomentosum} &  4,009\\       
        \textit{Caryocar brasiliensis}   & 19,639\\ 
        \textit{Myrcia guianensis}       &  3,233\\ 
        \textit{Miconia rubiginosa}      & 16,245\\ 
        \textit{Pouteria ramiflora}      &  4,284\\ 
        \textit{Pouteria torta}          & 12,552\\ 
        \hline
        Total                            & 59,962\\ 
        \bottomrule
    \end{tabular}
    \caption{Distribution of annotated pixels per class in the Itirapina dataset.}
    \label{tab:itirapina}
\end{table}

\subsection{Implementation Details}
In our experiments, we adopted a compact ViT architecture. 
The Transformer encoder consists of $L=6$ layers with 8 attention heads per layer.
The latent embedding dimension was set to $D=256$, and the internal MLP width was set to 512, utilizing GELU activation functions.
Our models were implemented in PyTorch (Python: 3.10.12, torch: 2.9.1, cuda: 12.5) and trained from scratch for 30 epochs using the AdamW optimizer with a learning rate of $3 \times 10^{-4}$ and weight decay of 0.01.
A dropout rate of 0.1 was applied to the attention weights and dense layers.

To ensure a rigorous evaluation, we benchmarked our ViT method against the state-of-the-art Multi-Temporal CNN~\cite{nogueira2019spatio}.
{For the Serra do Cipó dataset, we utilized the exact configuration and hyperparameters established by the original CNN baseline.}
However, the original study framed plant species identification on the Itirapina dataset as a one-against-all classification problem—training separate binary classifiers for each species—and used an older, less dense version of this benchmark. 
Consequently, a direct comparison with the originally reported results is not feasible.
To ensure a fair comparison regarding model capacity and multiclass performance, we adapted the Multi-Temporal CNN using the authors' official source code\footnote{\url{https://github.com/keillernogueira/spatio-temporal-phenological-segmentation/} (As of December 2025)}.
Because identifying species based on phenological changes is inherently a multi-class challenge, we modified the original implementation minimally, primarily replacing the final layer to support multi-class classification via softmax activation and cross-entropy loss.
{This decision ensures our evaluation more accurately reflects real-world deployment conditions while simultaneously subjecting the models to a more rigorous task.}
To guarantee the CNN baseline was not disadvantaged by this adaptation, we conducted a comprehensive grid search to re-optimize its key hyperparameters. However, no significant performance differences were observed across all configurations. For this reason, we retained the originally reported hyperparameters.

All experiments were conducted on a workstation equipped with an Intel Core i9-10900X 10-core CPU, 64 GB of DDR4 RAM, and a NVIDIA RTX 2080 Ti GPU.
The system runs Ubuntu 22.04.4 LTS with Linux kernel 5.15.0 and the ext4 file system.

\subsection{Performance Metrics}
Given the significant class imbalance observed in both the Serra do Cipó and Itirapina datasets (see Tables~\ref{tab:serra_do_cipo}~and~\ref{tab:itirapina}), standard Accuracy (Acc) can be misleading, as it is heavily biased toward dominant species.
Therefore, following the protocol established by Nogueira~et~al.~\cite{nogueira2019spatio}, we adopt \textbf{Balanced Accuracy (BAcc{.})} as the primary metric to assess classification performance.
Balanced accuracy is defined as the macro-average of recall (sensitivity) across all classes—calculating the arithmetic mean of the accuracy for each individual class.
This ensures that performance on rarer species contributes equally to the final score.
Consistent with Nogueira~et~al.~\cite{nogueira2019spatio}, evaluation is performed exclusively on annotated pixels, ignoring unclassified background regions.

To assess the computational efficiency and suitability for resource-constrained environments, we report:
\begin{itemize}
    \item \textbf{Parameters (M):} The total number of learnable weights in the model (in millions), representing the memory footprint and storage requirements.
    \item \textbf{Floating Point Operations (FLOPs):} The number of operations (in Gigas, GFLOPs) required to process a single input sample—i.e., one pixel's spatio-temporal sequence—during inference. This serves as a proxy for processing load and energy consumption. 
\end{itemize}
These values were measured using the THOP library\footnote{\url{https://pypi.org/project/thop/} (As of December 2025).}.

\section{Experimental Results}
\label{sec:results}

This section details the experimental evaluation of our ViT model. 
First, we conduct a comprehensive ablation study to identify the optimal ViT configuration for spatio-temporal vegetation pixel classification, systematically analyzing the design dimensions defined in Section~\ref{sec:proposed_methodologies}. 
Subsequently, {we conduct an analysis of the scalability of our method with bigger spatial square context window sizes.}
{Finally, }we compare the optimized ViT against the state-of-the-art Multi-Temporal CNN~\cite{nogueira2019spatio}, analyzing the trade-offs between classification performance and computational efficiency.

\subsection{Ablation Studies}

We conducted a two-stage ablation study to determine the optimal ViT configuration based on the design space defined in Section~\ref{sec:proposed_methodologies}. 
The first stage focused on the input data representation, evaluating data normalization, spectral arrangement, {spatial} context window shape, and tokenization strategy.
Building on these findings, we selected the most effective input configuration and proceeded to the second stage, where we assessed architectural components and boundary policies, specifically boundary handling, positional encoding, and feature aggregation mechanisms.

In both stages, experiments were performed only on the training splits of the Serra do Cipó and Itirapina datasets. 
For each dataset, a validation set was generated by randomly holding out 20\% of the training pixels. 
For each experiment, we report the best accuracy (Acc) achieved on this validation set, calculated as the ratio of correctly classified pixels to the total number of annotated pixels.

\subsubsection{Input Data Representation}
We first evaluated 24 combinations of input settings on both datasets to determine the optimal strategy to represent phenological data as tokens.
The validation results for the Serra do Cipó and Itirapina datasets are detailed in Tables~\ref{tab:vit_results_serra_cipo}~and~\ref{tab:vit_results_itirapina_v2}, respectively.

As evidenced in Table~\ref{tab:vit_results_serra_cipo}, the Serra do Cipó dataset exhibits clear performance saturation, with nearly all configurations achieving near-perfect accuracy ($>$99.9\%). 
This suggests that the phenological signatures of the four classes are sufficiently distinct to be captured by the ViT, regardless of the specific input representation. 
However, while these results validate the model's capacity, they offer little guidance for selecting specific hyperparameters.

\begin{table}[!htb]
\centering
\caption{Validation accuracy (best epoch) for different input configurations on Serra do Cipó dataset.}
\label{tab:vit_results_serra_cipo}
\setlength{\tabcolsep}{4pt}
\small
\begin{tabular}{c c l c c c}
\toprule
Setting & Norm. & Arr. & Window & Token & Acc. (\%) \\
\midrule
1  & No  & rgbrgb & Cross  & T,S & 100.00 \\
2  & No  & rgbrgb & Cross  & S,T & 100.00 \\
3  & No  & rgbrgb & Square & T,S & 100.00 \\
4  & No  & rgbrgb & Square & S,T & 100.00 \\
5  & No  & rrggbb & Cross  & T,S & 100.00 \\
6  & No  & rrggbb & Cross  & S,T & 100.00 \\
7  & No  & rrggbb & Square & T,S & 100.00 \\
8  & No  & rrggbb & Square & S,T & 100.00 \\
9  & Yes & rgbrgb & Cross  & T,S &  99.99 \\
10 & Yes & rgbrgb & Cross  & S,T &  99.93 \\
11 & Yes & rgbrgb & Square & T,S &  99.99 \\
12 & Yes & rgbrgb & Square & S,T &  99.96 \\
13 & Yes & rrggbb & Cross  & T,S &  99.98 \\
14 & Yes & rrggbb & Cross  & S,T &  99.94 \\
15 & Yes & rrggbb & Square & T,S &  99.98 \\
16 & Yes & rrggbb & Square & S,T &  99.97 \\
17 & No  & rgbrgb & Single & T,S & 100.00 \\
18 & No  & rgbrgb & Single & S,T & 100.00 \\
19 & No  & rrggbb & Single & T,S & 100.00 \\
20 & No  & rrggbb & Single & S,T & 100.00 \\
21 & Yes & rgbrgb & Single & T,S &  99.96 \\
22 & Yes & rgbrgb & Single & S,T &  99.94 \\
23 & Yes & rrggbb & Single & T,S &  99.96 \\
24 & Yes & rrggbb & Single & S,T &  99.96 \\
\bottomrule
\end{tabular}
\end{table}

\begin{figure}[!htb]
    \centering
    \includegraphics[width=0.45\linewidth]{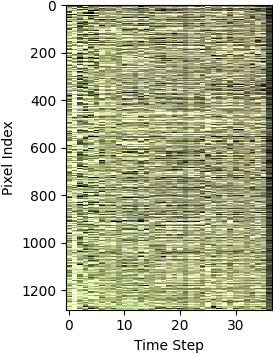}
    \includegraphics[width=0.45\linewidth]{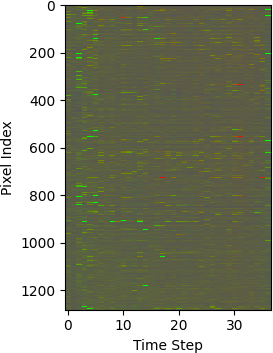}
    \caption{{Phenological visual rhythms for unnormalized (left) and normalized (right) pixels of an instance of the species with the best classification performance, \textit{Pouteria ramiflora}.}} 
    \label{fig:ramiflora}
\end{figure}

\begin{figure}[!htb]
    \centering
    \includegraphics[width=0.45\linewidth]{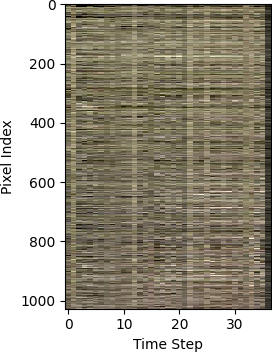}
    \includegraphics[width=0.45\linewidth]{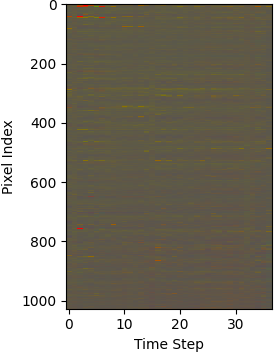}
    \caption{{Phenological visual rhythms for unnormalized (left) and normalized (right) pixels of an instance of the species with the worse classification performance, \textit{Miconia rubiginosa}.}}
    \label{fig:rubiginosa}
\end{figure}

The Itirapina dataset, with its daily temporal resolution and subtle inter-class variations, provided a more rigorous testing ground (see Table~\ref{tab:vit_results_itirapina_v2}).
Normalization generally failed to improve performance and, in several instances, caused significant degradation.
Configurations using raw data consistently yielded the highest validation scores, with Settings 4 (99.80\%), 17 (99.38\%), and 19 (99.23\%) leading the results.
Conversely, normalized settings exhibit both lower peak performance and greater variability; while Setting 15 achieved a competitive 97.75\%, other normalized configurations degraded significantly, dropping below 86\% (e.g., Settings 10, 12, 14, and 22).
This suggests that radiometric normalization interacts negatively with certain input configurations—particularly spatial Tokens (S,T) and single pixel context windows—and is therefore not a universally beneficial preprocessing step.
{To better illustrate this phenomenon, we generated visualizations of the phenological visual rhythms~\cite{almeida2016phenological}—plotting pixel intensity on the y-axis against timestamps on the x-axis. Figure~\ref{fig:ramiflora} show the visual rhythms for unnormalized and normalized pixels for one instance of the species with the best accuracy (\textit{Pouteria ramiflora}), and Figure~\ref{fig:rubiginosa} the species with the worst accuracy (\textit{Miconia rubiginosa}). We used the standard configuration we set on our paper.}

{By analyzing these spectral fingerprints, the detrimental effect of normalization becomes empirically clear. In dense tropical canopies, a species' unique signature is often defined by subtle, absolute changes in temporal intensity. As seen in the visualizations, normalization compresses the scale of these absolute reflectance differences. This forces the model to rely almost entirely on chromaticity, effectively ``washing out'' the subtle temporal intensity shifts that act as the discriminative features for these species.}

{As expected, by using a fully connected linear layer to implement the projection utilized early in our ViT architecture, the spectral organization of the data does not significantly impact the performance of our models.}

\begin{table}[!htb]
\centering
\caption{Validation accuracy (best epoch) for different input configurations on the Itirapina dataset.}
\label{tab:vit_results_itirapina_v2}
\setlength{\tabcolsep}{4pt}
\small
\begin{tabular}{c l l c l c}
\toprule
Setting & Norm. & Arr. & Window & Token & Acc. (\%) \\
\midrule
1  & No  & rgbrgb & Cross  & T,S & 99.22 \\
2  & No  & rgbrgb & Cross  & S,T & 98.78 \\
3  & No  & rgbrgb & Square & T,S & 98.72 \\
4  & No  & rgbrgb & Square & S,T & 99.80 \\
5  & No  & rrggbb & Cross  & T,S & 98.96 \\
6  & No  & rrggbb & Cross  & S,T & 98.77 \\
7  & No  & rrggbb & Square & T,S & 98.85 \\
8  & No  & rrggbb & Square & S,T & 99.58 \\
9  & Yes & rgbrgb & Cross  & T,S & 95.40 \\
10 & Yes & rgbrgb & Cross  & S,T & 81.76 \\
11 & Yes & rgbrgb & Square & T,S & 96.64 \\
12 & Yes & rgbrgb & Square & S,T & 85.54 \\
13 & Yes & rrggbb & Cross  & T,S & 95.59 \\
14 & Yes & rrggbb & Cross  & S,T & 82.23 \\
15 & Yes & rrggbb & Square & T,S & 97.75 \\
16 & Yes & rrggbb & Square & S,T & 85.44 \\
17 & No  & rgbrgb & Single & T,S & 99.38 \\
18 & No  & rgbrgb & Single & S,T & 94.58 \\
19 & No  & rrggbb & Single & T,S & 99.23 \\
20 & No  & rrggbb & Single & S,T & 95.37 \\
21 & Yes & rgbrgb & Single & T,S & 90.88 \\
22 & Yes & rgbrgb & Single & S,T & 78.09 \\
23 & Yes & rrggbb & Single & T,S & 90.76 \\
24 & Yes & rrggbb & Single & S,T & 79.07 \\
\bottomrule
\end{tabular}
\end{table}

Regarding spatial context, while single-pixel inputs performed surprisingly well on raw data (e.g., Setting 17: 99.38\%), they proved highly fragile under normalization, with accuracy dropping as low as 79.07\% (Setting 24). 
In contrast, the cross-context window with 5 pixels and the square context window with 3$\times$3 pixels configurations remained comparatively stable even when normalized (e.g., Settings 9, 13, and 15 all maintained $>$95\% accuracy). 
This suggests that spatial context helps mitigate the potential loss of radiometric information caused by normalization.

Finally, temporal tokens (T,S) proved generally more stable and often superior to spatial tokens (S,T), particularly in normalized regimes.
For instance, under normalization with an 3$\times$3 pixels square context window, Setting 11 (T,S) reached 96.64\%, whereas its spatial counterpart, Setting 12 (S,T), dropped to 85.54\%.
This aligns with the biological intuition that prioritizing temporal continuity—allowing self-attention to model dependencies between dates—is advantageous for capturing phenological dynamics.

Based on these findings, we established the following baseline configuration for the second stage: (\textit{i}) Raw Data; (\textit{ii}) Natural Order (rgbrgb); (\textit{iii}) square context window; and (\textit{iv}) Temporal Token (T, S).
We selected the square context window as it offers the optimal stability (superior to single-pixel) and more context with minimal increase in computational complexity (similar amount of parameters and FLOPs to the cross-context window).
Regarding the spectral arrangement, the ablation revealed no significant difference between natural order (rgbrgb) and channel grouping (rrggbb); therefore, the former was adopted to minimize preprocessing overhead.
Given that the Serra do Cipó dataset exhibited minimal sensitivity to these variations, we adopted this robust configuration for both datasets.

\subsubsection{Architectural Components and Boundary Policies}

After selecting the best configuration for the input data, we focused our analysis on architectural aspects of the model. We evaluate approaches for handling neighboring pixels outside the annotated region, alongside architectural variations regarding positional encoding and feature aggregation strategies. 
The results for both the Serra do Cipó and Itirapina datasets are presented in Table~\ref{tab:ablation_best_only}.

\begin{table}[!htb]
\centering
\caption{Validation accuracy (best epoch) for different boundaries policies and architectural variation on Serra do Cipó and Itiratina datasets.}
\label{tab:ablation_best_only}
\setlength{\tabcolsep}{2pt}
\small
\begin{tabular}{l l c l c c l c}
\toprule
Norm. & Arr. & window & Token & Bound. & PosEnc & Agg. & Acc. (\%) \\
\midrule
\multicolumn{8}{l}{\textbf{Serra do Cipó}} \\
\midrule
No  & rgbrgb & Square & T,S & Black & Yes & CLS & 100.00 \\
No  & rgbrgb & Square & T,S & Real  & Yes & CLS & 100.00 \\
No  & rgbrgb & Square & T,S & Black & No  & CLS &  99.99 \\
No  & rgbrgb & Square & T,S & Black & Yes & GAP & 100.00 \\
\midrule
\multicolumn{8}{l}{\textbf{Itirapina}} \\
\midrule
No  & rgbrgb & Square & T,S & Black & Yes & CLS &  98.52 \\
No  & rgbrgb & Square & T,S & Real  & Yes & CLS &  98.69 \\
No  & rgbrgb & Square & T,S & Black & No  & CLS &  97.37 \\
No  & rgbrgb & Square & T,S & Black & Yes & GAP &  97.86 \\
\bottomrule
\end{tabular}
\end{table}

For the Serra do Cipó dataset, validation accuracies  were again extremely high ($\geq$ 99.99\%), with three configurations reaching a perfect score of 100.00\%.
The performance differences between boundary handling approaches (Black Padding vs. Real Value), positional encoding settings (Yes vs. No), and feature aggregation strategies ([CLS] vs. GAP) were negligible.
This confirms that the classification task on this dataset is performance-saturated under the current protocol. 
Consequently, the Serra do Cipó results offer limited insight into specific preferences, as the model effectively solves the problem regardless of the tested design decisions.

In contrast, the Itirapina dataset exhibited a {slightly} wider spread of results (97.37 to 98.69), providing more actionable insights. 
Regarding boundary handling, using the real values of external pixels yielded a slight improvement over black padding (98.69\% vs. 98.52\%), suggesting that preserving radiometric continuity provides useful spatial context. 
However, the most significant factors were architectural: disabling positional encoding caused a performance drop from 98.52\% to 97.37\%, confirming the importance of learning the sequence order of tokens.
Moreover, the [CLS] token outperformed Global Average Pooling (98.52\% vs. 97.86\%), indicating that a learned token better summarizes the input sequence than simple averaging.

Based on these findings, we selected the final configuration for our ViT model: (\textit{i}) Black Padding; (\textit{ii}) With Positional Encoding; and (\textit{iii}) Class Token ([CLS]).
Although using real values for boundary handling yielded a marginal accuracy gain on Itirapina ($<$ 0.17\%), black padding was chosen because it introduces input sparsity, which can be exploited for computational efficiency.

\subsection{{Scalability Analysis}}
\label{sec:scalability}

\begin{figure}[!htb]
    \centering
    
    \includegraphics[width=\linewidth]{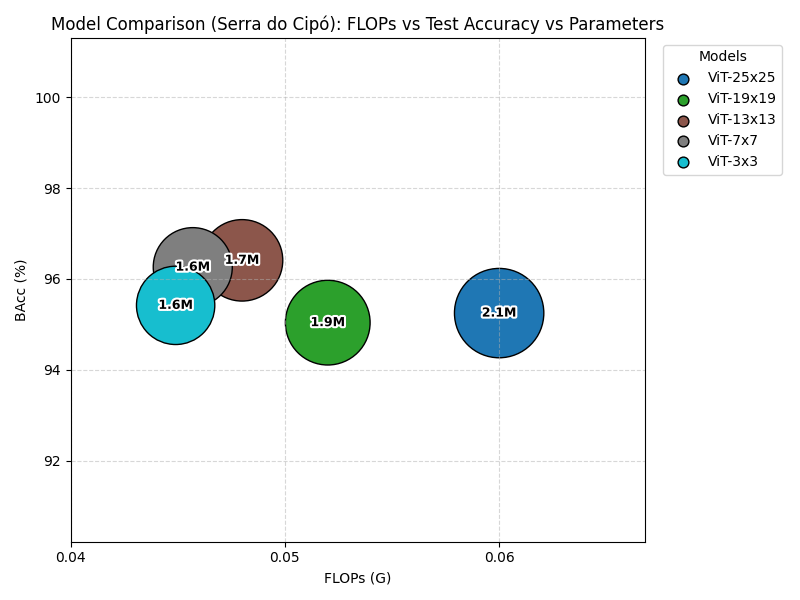}\\
    (a)
    
    \includegraphics[width=\linewidth]{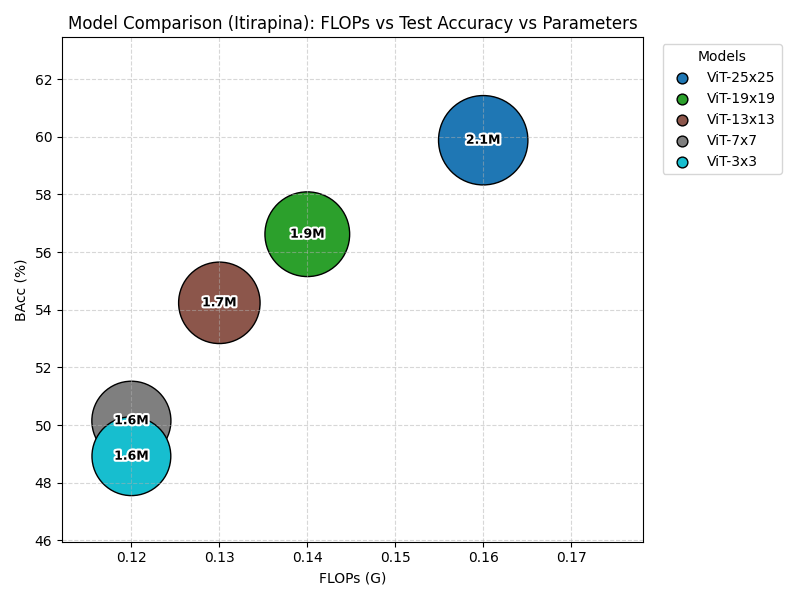}\\
    (b)
    
    \caption{{Balanced accuracy, computational complexity (FLOPs), and number of parameters for our ViT on the (a) Serra do Cipó and (b) Itirapina datasets with square context window sizes varying from 3x3 to 25x25 pixels. The number of parameters is represented by the diameter of the circles and is written inside them.}} 
    \label{fig:AccFlops}
\end{figure}

{After configuring our ViT model, we investigated the scalability of our method by varying the size of the square context window to further understand the influence of spatial context.}
{Specifically, we evaluated square context window sizes of 3$\times$3, 7$\times$7, 13$\times$13, 19$\times$19, and 25$\times$25 pixels. The objective was to map the trade-offs between classification performance (balanced accuracy) on the test set, computational complexity (FLOPs), and number of parameters. We report the average result for three runs of each experiment. Figure~\ref{fig:AccFlops} illustrate this trade-off for the Serra do Cipó and Itirapina datasets.}

{The datasets exhibited different behaviors to the expansion in spatial context due to their distinct acquisition modalities and spatial resolutions.}
{For the Serra do Cipó dataset (aerial UAV imagery), increasing the square context window beyond a certain threshold proved detrimental. As seen in Figure~\ref{fig:AccFlops} (a), the optimal balance was achieved with smaller windows, peaking at 13$\times$13 pixels, which reached a balanced accuracy of 96.41\%, 1.72M parameters and 0.05 GFLOPs. As the window size expanded, accuracy degraded, culminating in 95.25\% for 25$\times$25, while the computational cost and parameter count increased, reaching  2.07M parameters and 0.06 GFLOPs for 25$\times$25.}
{This behavior indicates that in aerial images, that are characterized by a high density of information per pixel, excessively large square context windows may inadvertently encompass multiple canopies.}
{This may introduce noise from different individuals or species, which may lead to less discriminative features.}

{Conversely, the Itirapina dataset (near-surface tower imagery) demonstrated a positive correlation between square context window size and balanced accuracy.}
{Figure~\ref{fig:AccFlops} (b) shows that the model's balanced accuracy scales almost linearly with the input size. The 3$\times$3 configuration yielded an balanced accuracy of approximately 48.92\%, with 1.59M parameters and 0.12 GFLOPs. Expanding the window progressively improved performance, culminating in the 25$\times$25 configuration, which achieved the highest balanced accuracy of 59.88\%, with  2.07M parameters and 0.16 GFLOPs. Because tower-based cameras capture finer, more localized details (e.g., individual leaves or branches), a single pixel contains less spatial information. A larger spatial window is therefore required to aggregate enough contextual data to properly characterize the canopy and identify the species.}

{These results highlight the flexibility of the ViT architecture. Unlike structurally rigid models, the ViT can be effortlessly scaled by adjusting the input size to meet user-specific needs, offering a tunable trade-off between maximizing balanced accuracy and prioritizing computational efficiency for deployment on resource-constrained edge devices.}

{For the next experiments, we set the square context window size to 13$\times$13 for the Serra do Cipó dataset, and 25$\times$25 for the Itirapina dataset, as these were the configurations that obtained the best balanced accuracy.}

\subsection{Comparison with Other Methods}
\label{sec:results:comparison}

Table~\ref{tab:cost_comparison} highlights a clear trade-off between model capacity, computational cost, and predictive performance across the two datasets. 
For both cases, our ViT was trained using the optimized configuration identified in the ablation study: (\textit{i}) Raw Data; (\textit{ii}) Natural Order (rgbrgb); (\textit{iii}) Black Padding; (\textit{iv}) square context window; and (\textit{v}) Temporal Token (T, S) (\textit{vi}) With Positional Encoding; and (\textit{vii}) Class Token ([CLS]).
{We also use the optimal square context window sizes identified in Section~\ref{sec:scalability}: a 13$\times$13 window for the Serra do Cipó dataset and a 25$\times$25 window for the Itirapina dataset.}
{From the three runs of each experiment we conducted, we report the balanced accuracy of the best one.} 

\begin{table*}[!htb]
\centering
\caption{Comparison of performance on the Serra do Cipó and Itirapina datasets.}
\label{tab:cost_comparison}
\setlength{\tabcolsep}{2pt}
\small
\begin{tabular}{c c c c}
\toprule
Method & Params (M) & FLOPs (G) & BAcc. (\%) \\
\midrule
\multicolumn{4}{l}{\textbf{Serra do Cipó}} \\
\midrule
Multi-temporal CNN~\cite{nogueira2019spatio} & 3.84 & 0.73 & 98.90 \\
ViT (ours)                                   & 1.72 & 0.05 & 96.79 \\
\midrule
\multicolumn{4}{l}{\textbf{Itirapina}} \\
\midrule
Multi-temporal CNN~\cite{nogueira2019spatio} & 7.95 & 2.07 & 53.62 \\
ViT (ours)                                   & 2.07 & 0.16 & 61.51 \\
\bottomrule
\end{tabular}
\end{table*}

\begin{table*}[!htb]
\centering
\caption{Confusion matrix for the Multi-temporal CNN~\cite{nogueira2019spatio} on the Serra do Cipó dataset.}
\label{tab:confusion_convnet_4c}
\setlength{\tabcolsep}{4pt}
\small
\begin{tabular}{l|rrrr}
\toprule
\textbf{GT $\backslash$ Pred} 
& \textit{B. virgilioides} 
& \textit{E. erythropappus} 
& \textit{V. cinnamomea} 
& {Evergreen species} \\
\midrule
\textit{B. virgilioides}             & 100.00\% & 0.00\% & 0.00\% & 0.00\% \\
\textit{E. erythropappus}            & 0.00\% & 100.00\% & 0.00\% & 0.00\% \\
\textit{V. cinnamomea}               & 0.00\% & 0.00\% & 99.14\% & 0.86\% \\
Evergreen species                    & 1.92\% & 0.44\% & 1.20\% & 96.44\% \\
\bottomrule
\end{tabular}
\end{table*}

\begin{table*}[!htb]
\centering
\caption{Confusion matrix for our Vision Transformer (ViT) model on the Serra do Cipó dataset.}
\label{tab:confusion_vit_4c}
\setlength{\tabcolsep}{4pt}
\small
\begin{tabular}{l|rrrr}
\toprule
\textbf{GT $\backslash$ Pred} 
& \textit{B. virgilioides} 
& \textit{E. erythropappus} 
& \textit{V. cinnamomea} 
& {Evergreen species} \\
\midrule

\textit{B. virgilioides}             & {93.33}\% &  {0.00}\% &  {3.20}\% &  {3.47}\% \\
\textit{E. erythropappus}            &  {0.24}\% & {99.37}\% &  {0.13}\% &  {0.26}\% \\
\textit{V. cinnamomea}               &  {2.14}\% &  {0.04}\% & {95.93}\% &  {1.89}\% \\
Evergreen species                    &  {0.00}\% &  {0.31}\% &  {1.17}\% & {98.52}\% \\

\bottomrule
\end{tabular}
\end{table*}

On the Serra do Cipó dataset, the proposed ViT achieves a competitive balanced accuracy of 96.79\% (trailing the baseline by only 2.11\%) while using 55\% fewer parameters (1.72M vs. 3.84M) and 14$\times$ fewer FLOPs (0.05G vs.\ 0.73G) compared to the Multi-Temporal CNN. 
Notably, our ViT offers a more favorable accuracy–efficiency ratio, achieving near-parity performance with a fraction of the computational resources.

The confusion matrices in Tables~\ref{tab:confusion_convnet_4c}~and~\ref{tab:confusion_vit_4c} reveal markedly different error profiles. 
The Multi-temporal CNN exhibits strong diagonal dominance, achieving perfect classification for \textit{B. virgilioides} and \textit{E. erythropappus}, with only minor confusion between \textit{V. cinnamomea} and the Evergreen species ($<$1\%), and between Evergreen species and other classes (3.56\%).

In contrast, our ViT shows a slightly more dispersed error distribution, indicating that while it captures general phenological trends, it struggles with the fine-grained separation of spectrally similar classes. Specifically, we improved the accuracy in the Evergreen species (98.52\%), which was the species that the CNN baseline had the lowest performance. Our ViT mainly confused Evergreen species with \textit{V. cinnamomea} (1.17\%). Our model also kept a similar accuracy for the \textit{E. erythropappus} species (99.37\%), with only a few examples being misclassified as other species ($<$0.63\%). On the other hand, it exhibits lower accuracy for \textit{B. virgilioides} (93.33\%) and \textit{V.} (95.93\%). Notably, 2.14\% of \textit{V. cinnamomea} pixels were misclassified as \textit{B. virgilioides} and 1.89\% as Evergreen species. For \textit{B. virgilioides}, 3.20\% of pixels were misclassified as \textit{V. cinnamomea} and 3.47\% as Evergreen species.
\begin{table*}[!htb]
\centering
\caption{Confusion matrix for the Multi-temporal CNN~\cite{nogueira2019spatio} on the Itirapina dataset.}
\label{tab:confusion_convnet}
\setlength{\tabcolsep}{6pt}
\small
\begin{tabular}{l|rrrrrr}
\toprule
\textbf{GT $\backslash$ Pred} & \textit{A. tomentosum} & \textit{C. brasiliensis} & \textit{M. guianensis} & \textit{M. rubiginosa} & \textit{P. ramiflora} & \textit{P. torta} \\
\midrule
\textit{A. tomentosum}   & 0.00\% & 91.95\% & 0.00\% & 0.00\% & 8.05\% & 0.00\% \\
\textit{C. brasiliensis} & 0.00\% & 80.20\% & 0.86\% & 18.19\% & 0.47\% & 0.27\% \\
\textit{M. guianensis}   & 0.00\% & 10.48\% & 41.52\% & 48.00\% & 0.00\% & 0.00\% \\
\textit{M. rubiginosa}   & 0.00\% & 0.00\% & 0.00\% & 100.00\% & 0.00\% & 0.00\% \\
\textit{P. ramiflora}    & 0.00\% & 0.00\% & 0.00\% & 0.00\% & 100.00\% & 0.00\% \\
\textit{P. torta}        & 0.00\% & 0.00\% & 0.00\% & 1.93\% & 98.07\% & 0.00\% \\
\bottomrule
\end{tabular}
\end{table*}

\begin{table*}[!htb]
\centering
\caption{Confusion matrix for our Vision Transformer (ViT) model on the Itirapina dataset.}
\label{tab:confusion_vit}
\setlength{\tabcolsep}{6pt}
\small
\begin{tabular}{l|rrrrrr}
\toprule
\textbf{GT $\backslash$ Pred} & \textit{A. tomentosum} & \textit{C. brasiliensis} & \textit{M. guianensis} & \textit{M. rubiginosa} & \textit{P. ramiflora} & \textit{P. torta} \\
\midrule
\textit{A. tomentosum}   & {59.33}\% &  {8.72}\% &  {0.00}\% &  {0.08}\% & {0.00}\% & {31.87}\% \\
\textit{C. brasiliensis} &  {0.04}\% & {75.80}\% &  {0.55}\% & {15.95}\% & {5.70}\% &  {1.96}\% \\
\textit{M. guianensis}   &  {0.00}\% &  {3.01}\% & {53.70}\% & {29.93}\% & {1.18}\% & {12.18}\% \\
\textit{M. rubiginosa}   &  {0.00}\% & {11.11}\% &  {0.77}\% & {83.87}\% & {0.00}\% &  {4.25}\% \\
\textit{P. ramiflora}    &  {0.00}\% & {81.51}\% &  {0.87}\% &  {0.90}\% & {8.03}\% &  {8.70}\% \\
\textit{P. torta}        &  {0.00}\% &  {4.87}\% &  {0.12}\% &  {0.32}\% & {6.38}\% & {88.31}\% \\

\bottomrule
\end{tabular}
\end{table*}

The advantages of our ViT are most pronounced on the Itirapina dataset.
In terms of classification performance, our ViT yielded a balanced accuracy of 61.51\%, surpassing the baseline by 7.89\%.
Notably, this result was achieved with over 3$\times$ fewer parameters (2.07M vs.\ 7.95M) and over 12$\times$ fewer FLOPs (0.16G vs.\ 2.07G) compared to the Multi-Temporal CNN. 
These results show that our ViT improved over the previous state-of-the-art while delivering massive efficiency gains.

The confusion matrix for Itirapina (see Table~\ref{tab:confusion_convnet}) shows that the Multi-Temporal CNN exhibits a strong bias toward dominant and visually distinctive species such as \textit{M. rubiginosa} and \textit{P. ramiflora}, achieving near-perfect performance on these classes while {completely} failing to identify rarer or subtler species like \textit{A. tomentosum} and \textit{P. torta}.
This suggests that the CNN relies heavily on discriminative local patterns and may overfit to majority-class characteristics. 

In contrast, our ViT demonstrates a more balanced performance across difficult classes (see Table~\ref{tab:confusion_vit}). 
Unlike the CNN, the ViT successfully recovers a significant portion of the \textit{A. tomentosum} (59.33\%) and \textit{P. torta} (88.31\%) instances.
{Our model mainly misclassified \textit{A. tomentosum} pixels with \textit{P. torta} (31.87\%) and \textit{C. brasiliensis} (8.72\%), while  \textit{P. torta} instances were primarily misclassified with \textit{P. ramiflora} (6.38\%) and \textit{C. brasiliensis} (4.87\%).}
{The accuracy for \textit{M. guianensis} also improved considerably when compared with the CNN baseline, increasing from 41.52\% to 53.70\%. The main sources of confusion for our ViT with this species were \textit{M. rubiginosa} (29.93\%) and \textit{P. torta} (12.18\%).}

However, these improvements in sensitivity came at the cost of precision {in some species.}
For the \textit{C. brasiliensis} species, our ViT reached 75.80\% accuracy, droping 4.4\% from the baseline, mainly confusing with the \textit{M. rubiginosa} (15.95\%) and the \textit{P. ramiflora} (5.70\%) species. For the \textit{M. rubiginosa} species, there was a drop of 16.13\% (83.87\% vs. 100\%), with 11.11\% of samples being misclassified as \textit{C. brasiliensis} and 4.25\% as \textit{P. torta}. The hardest species to classify for our ViT model was the P. ramiflora, achieving only 8.03\% accuracy. The main factor to this low performance was the confusion with the C. brasiliensis, with the majority (81.51\%) of P. ramiflora pixels being misclassified as this species.

\section{{Discussions}}
\label{sec:discussions}

{In this section, we discuss the implications of our findings, our current limitations, and outline potential avenues for future research to further refine ViTs for phenological monitoring.}

\subsection{Generalization for Other Biomes and Data Modalities}
\label{sec:biomes_modalities}

{The primary focus of this study was to address the unique challenges presented by the dense tropical environments of the Brazilian Cerrado. This biome is characterized by an extreme diversity of species, overlapping canopies, and highly subtle phenological variations, making pixel-wise classification particularly difficult~\cite{ECOI_2014_Alberton}.
While our optimized ViT demonstrated significant efficiency and competitive accuracy on both aerial and near-surface tower imagery within this context, caution must be exercised when generalizing these findings.}
{Due to the stark differences in spatial resolution between these modalities, the temporal and spatial signature of pixels can vary significantly. Furthermore, evaluating the proposed architecture across different biomes, such as temperate forests with more pronounced seasonal changes, or on satellite imagery remains a necessary next step.}

{Recent studies in self-supervised learning, such as AgriFM~\cite{li2025agrifm}, suggest that foundational models can generate robust, multi-scale spatio-temporal representations. However, these approaches require massive amounts of annotated data.}
{While we currently have a large amount of pixels to train our models, there are still a small number of annotated instances and few different images (timestamps). Future work will explore how foundational models can be leveraged or adapted to resource-constrained phenological datasets to achieve broader generalization.}

\subsection{Input Data Order and Multi-branch Projection Layers}
\label{sec:data_order}

Our ablation study regarding spectral arrangement revealed no significant performance difference between preserving the natural temporal sequence (rgbrgb) and grouping channels color (rrggbb). This behavior is expected due to the projection utilized early in the ViT architecture being implemented using a linear layer.

However, the effects would be significantly more pronounced if alternative projection architectures were employed. For example, a multi-branch projection layer would correlate single-channel data before mixing spectral data from the entire input. In it, exclusive Multilayer Perceptrons (MLPs) are assigned to each specific branch before fusing the embeddings. Each branch handle a group of channels, this way, the initial order of the inputs is more significant since it defines how these groups are constructed. We anticipate that a more structured feature extraction phase like this one would be especially critical when incorporating multi-spectral and hyperspectral data.

\subsection{Spatio-Temporal 3D Tokens}
\label{sec:3d_tokens}

To maintain architectural simplicity and limit the scope of our hyperparameter search space in this initial study, since we already analyze several other dimensions of possible ViT modifications, our framework utilized 1D sequence flattening strategies (exclusively temporal or spatial tokens).
However, the use of a hybrid 3D tokenization strategy, such as utilizing 3D patches combined with a 3D convolutional layer for embedding, or employing multi-dimensional attention, is a highly promising approach for capturing joint spatio-temporal dynamics.
Recent architectures, such as TSViT~\cite{tarasiou2023vits}, have explored factorized spatio-temporal attention to handle time series data effectively in satellite imagery. While integrating 3D tokens would substantially expand the experimental complexity, it represents a highly relevant direction for future architectures, particularly as the methodology scales to accommodate more complex input data, such as multi-spectral and hyperspectral images.

\subsection{Multi-spectral and Hyperspectral Images}
\label{sec:multi_hyper_spectral}

The current iteration of our proposed method focuses exclusively on RGB images. However, remote sensing and ecological monitoring frequently rely on multi-spectral and HyperSpectral Imaging (HSI) to capture richer, more discriminative spectral fingerprints beyond the visible light spectrum. Transitioning our efficient ViT to handle hyperspectral data will introduce new dimensional challenges. Recent advancements in the field offer promising solutions, such as HyPyraMamba~\cite{li2026hypyramamba}, which utilizes a Pyramid Spectral Attention (PSA) module alongside a Mamba module for robust sequence modeling to overcome standard transformer limitations. Additionally, techniques like SWDiff~\cite{chen2024swdiff} have demonstrated that diffusion models can effectively augment hyperspectral datasets to improve classification performance. Adapting our ViT framework to process multi-spectral and hyperspectral data using similar strategies is a primary objective for future work.

\section{Conclusion}
\label{sec:conclusion}
In this work, we presented a comprehensive study on optimizing ViT for spatio-temporal vegetation pixel classification in phenological time series. 
Motivated by the structural rigidity and computational cost of state-of-the-art approaches, such as the Multi-Temporal CNN~\cite{nogueira2019spatio}, we explored the viability of ViTs as a robust, scalable alternative for monitoring plant functional traits.
We evaluated our models on two diverse datasets from the Brazilian Cerrado biome: Serra do Cipó (aerial imagery) and Itirapina (near-surface imagery).

Our extensive ablation study, analyzing seven key design dimensions, revealed that the optimal deployment of ViTs depends heavily on dataset complexity.
While the Serra do Cipó dataset proved largely insensitive to configuration changes due to performance saturation, the Itirapina dataset was highly sensitive to design choices.
Based on these findings, we identified a robust input configuration combining raw (unnormalized) data in their natural order (rgbrgb), a square context {window}, and temporal tokens.
Additionally, employing black padding for boundary handling, learnable positional encoding, and a class token ([CLS]) for feature aggregation yielded the optimal balance between performance and stability.
{For aerial imagery, where there is a higher density of information, smaller square context window sizes such as 13$\times$13 pixels obtained better results, while for tower-based images, where a single pixel contain more granular details, bigger square context windows were favored, such as 25$\times$25 pixels.}
The most significant contribution of this study is the demonstration of superior computational efficiency.
Our optimized ViT achieves competitive {or better} classification performance while decreasing Floating Point Operations (FLOPs) by an order of magnitude and maintaining constant parameter complexity regardless of the time series length, whereas the baseline CNN scales linearly.
This massive reduction in computational overhead makes our ViT framework feasible for phenological monitoring on resource-constrained edge devices.

Additionally, our comparative analysis highlighted a distinct trade-off in classification behavior.
While the state-of-the-art CNN achieved higher accuracy on dominant and visually distinctive species, it frequently failed to identify rarer or subtler species.
In contrast, our ViT approach, despite a slightly lower overall accuracy, exhibited significantly better sensitivity to difficult classes, such as \textit{Aspidosperma tomentosum} and \textit{Pouteria torta} in the Itirapina dataset.
This suggests that the ViT's self-attention mechanism is more effective at capturing the subtle temporal cues of species that lack distinct spatial features.

For future works, we plan to: expand this analysis to include hierarchical architectures, such as Swin Transformers, to better capture multi-scale spatial features;
{adapt our method to multi-spectral and hyperspectral data, following strategies like SWDiff~\cite{chen2024swdiff} and HyPyraMamba~\cite{li2026hypyramamba}; use multi-branch projection layers with a MLP exclusive for each branch; experiment with 3D tokenization strategies;}
and explore {the} open-set recognition scenarios, further refining our ViT for open-world environmental monitoring, where unknown or invasive species may be present.

\section*{Acknowledgments}
This research was supported by São Paulo Research Foundation - FAPESP (\#2023/17577-0, \#2024/04500-2, \#2024/22985-3) and National Council for Scientific and Technological Development - CNPq (\#315220/2023-6, \#420442/2023-5, \#444982/2024-8).

\bibliographystyle{IEEEtran}
\bibliography{refs}

\end{document}